\documentclass[runningheads]{llncs}
\usepackage[T1]{fontenc}
\usepackage{amsfonts}
\usepackage{amsmath}
\usepackage{bm}
\usepackage{graphicx,verbatim}
\usepackage{cite}
\usepackage{hyperref}
\usepackage{xurl}
\usepackage{booktabs}
\usepackage{orcidlink}

\usepackage{color}

\begin{document}
\title{Modelling Geographic Atrophy Progression using Implicit Neural Representations}
\titlerunning{Modelling GA Progression using INRs}
%

\author{Simone Sarrocco\inst{1,2}\orcidlink{0009-0007-5854-7382} \and
Paul Friedrich\inst{1}\orcidlink{0000-0003-3653-5624} \and
Florentin Bieder\inst{1}\orcidlink{0000-0001-9558-0623} \and
Christina Bornberg\inst{4}\orcidlink{0000-0002-1860-2005} \and
Philippe Valmaggia\inst{1,2,3}\orcidlink{0000-0003-2817-5630} \and
Peter M. Maloca\inst{1,2,3,*}\orcidlink{0000-0002-4794-5859} \and
Philippe C. Cattin\inst{1,*}\orcidlink{0000-0001-8785-2713}}

\authorrunning{S. Sarrocco et al.}
%
\institute{Department of Biomedical Engineering, University of Basel, Allschwil, Switzerland 
    \email{simone.sarrocco@unibas.ch}\and
Department of Ophthalmology, University Hospital Basel, Basel, Switzerland\\\and
Moorfields Eye Hospital NHS Foundation Trust, London, United Kingdom\\\and Vienna University of Natural Resources and Life Sciences, Vienna, Austria\\ *Shared last authorship}

\maketitle              
\begin{abstract}
Age-related Macular Degeneration (AMD) is the major cause of blindness in the Western world. Its late dry phase is characterised by irreversible atrophic areas, namely Geographic Atrophy (GA). Longitudinal Fundus Autofluorescence (FAF) image acquisitions are currently the main tool for assessing lesion growth over time at the image level. However, due to its highly individualised progression, the evolution of late AMD remains poorly understood. In this work, we propose using Implicit Neural Representations (INRs) to model GA progression at the individual level in a low-data setting. Our approach generates both FAF and GA segmentation at both past and future time points. Among the comparison models, our method achieves competitive segmentation quality across different scenarios, yielding the lowest Mean Absolute Error (MAE) for the GA lesion area and the highest DICE score, without sacrificing FAF image quality. The code is available at~\mbox{\url{https://github.com/SimoneSarrocco/ga-progression-with-inrs}}.

\keywords{Generative Modelling  \and Age-related Macular Degeneration \and Disease Progression}

\end{abstract}

\section{Introduction}
Age-related macular degeneration (AMD) is a progressive disease of the macula and one of the leading causes of blindness in individuals over 55, affecting nearly 9\% of the global population~\cite{fleckensteinAgeRelatedMacularDegeneration2024,wongGlobalPrevalenceAgerelated2014a}. While early stages may be asymptomatic, late-stage AMD can severely impair vision and everyday activities such as reading, driving, and recognising faces~\cite{fleckensteinProgressionGeographicAtrophy2018}. 
Its dry form, characterised by the development of geographic atrophy (GA), still lacks robust treatment, underscoring the need for novel therapies~\cite{fleckensteinAgerelatedMacularDegeneration2021}. Fundus autofluorescence (FAF) has become the main modality for measuring GA progression~\cite{poleFundusAutofluorescenceClinical2021}, and understanding this progression within and between individuals is crucial for designing clinical trials and informing patients about prognosis~\cite{wongGlobalPrevalenceAgerelated2014a}.

In this work, we model subject-specific GA progression in patients with late dry AMD despite limited longitudinal data. Our modelling assumption is that the GA progression follows a continuous, patient-specific trajectory, with FAF images representing observations at discrete time points. Using Implicit Neural Representations (INRs), we jointly learn lesion growth trajectories from FAF images and corresponding GA segmentation masks. By adapting a population-level disease trajectory to subject-specific data, we model the disease progression as a continuous process, enabling interpolation of intermediate states and prediction of future GA progression at both image and segmentation levels.

\begin{figure}[t]
    \centering
    \includegraphics[width=\textwidth]{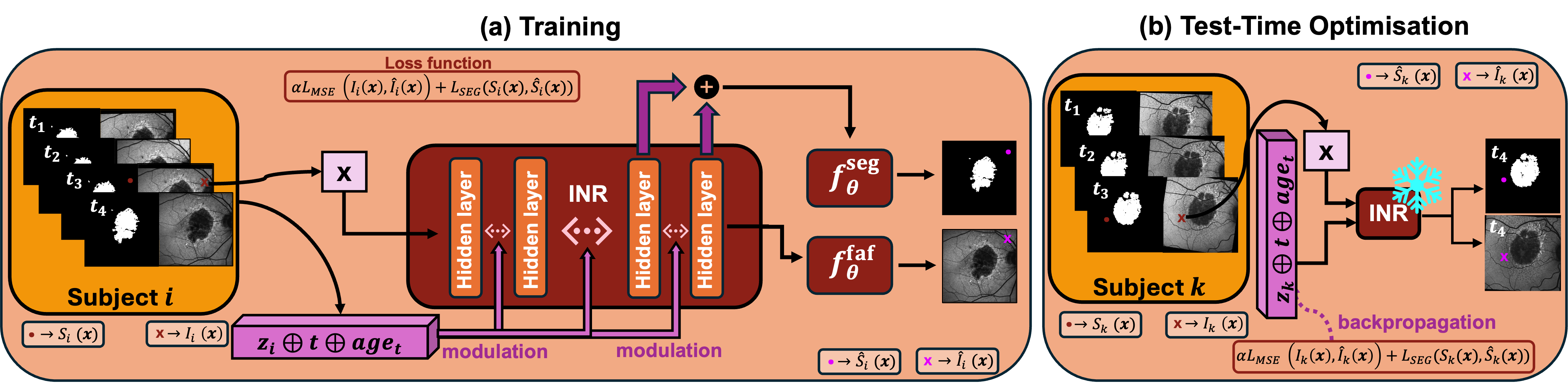}
    \caption{Our proposed model architecture. In the training phase (a), a spatial coordinate $\mathbf{x}\in\mathbb{R}^2$ is fed into the MLP. Each layer of the MLP is conditioned by concatenating the latent vector $z_i$, the time in weeks from the baseline visit, $t$, and the age of the patient at $t$, $\text{age}_t$, via modulation. The shared INR is split into reconstruction head $f_\theta^{\text{faf}}$, which outputs the pixel intensity value $\hat{I}_i(\mathbf{x})$ (purple cross in the output FAF), and the segmentation head $f_\theta^{\text{seg}}$, which takes the outputs from the last two hidden layers and outputs the GA segmentation label $\hat{S}_i(\mathbf{x})$ (purple point in the output segmentation). The predictions are compared to the ground truth $I_i(\mathbf{x})$ (red cross in the input FAF) and $S_i(\mathbf{x})$ (red point in the input segmentation), respectively, using a combination of MSE ($\mathcal{L}_{\text{MSE}}$, for FAF reconstruction) and sum of DICE and Binary Cross Entropy ($\mathcal{L}_\text{SEG}$, for GA segmentation) as loss function. During test-time optimisation (b), a new latent vector $z_k$ for a new eye $k$ is randomly sampled from $\mathcal{N}(0,10^{-2})$ and optimised across both FAF images and GA segmentations using the same loss function, while keeping the INR parameters frozen. By concatenating the new optimised latent vector with the time-conditioning variables, the forward pass generates the new FAF image and GA segmentation at the specified time.}
    \label{fig:model_architecture}
\end{figure}

\paragraph{\textbf{Related work}.} Deep learning (DL) approaches have become popular for modelling disease progression due to their ability to learn spatio-temporal changes directly from medical scans. To model GA progression, Lachinov et al.~\cite{lachinovLearningSpatioTemporalModel2024a} developed a pixel-level disease progression model based on Neural Ordinary Differential Equations (NODEs) that predicts future lesion segmentations from Optical Coherence Tomography (OCT) en-face projections. Salvi et al.~\cite{salviDeepLearningPredict2025b} trained different U-Net architectures on FAF images to predict the region of growth of GA. Liu et al.~\cite{liuImageFlowNetForecastingMultiscale2025d} introduced ImageFlowNet, a U-Net architecture incorporating NODEs for modelling pixel-level changes in GA lesions on FAF images. More recently, INRs have emerged as a flexible framework for learning continuous representations of complex data. Within medical imaging, INRs have been adopted for many different applications, including image overfitting~\cite{friedrichMedFunctaUnifiedFramework2026a}, segmentation~\cite{stolt-ansoNISFNeuralImplicit2023,vyasFitPixelsGet2026}, atlas generation~\cite{danneckerCINeMAConditionalImplicit2026}, and longitudinal image registration~\cite{shuaibuCapturingLongitudinalChanges2025}. 
Lately, INRs have also been utilised for disease progression modelling. Bieder et al.~\cite{biederModelingNeonatalBrain2025} employed INRs to model neonatal brain development by learning a continuous age-agnostic representation of the identity of the subject. Dannecker et al.~\cite{danneckerPredictingLongitudinalBrain2026} predicted individual past and future trajectories over a time span of up to 20\,weeks from single MRI scans, conditioned on external variables.
\paragraph{\textbf{Contribution}.} To the best of our knowledge, this work presents the first application of INRs to model individual lesion trajectories in GA secondary to AMD. Our method predicts FAF images and GA segmentation masks over time while providing accurate estimates of lesion size at each time point. In a low-data setting, the model generates past and future segmentations from a single combination of FAF image and corresponding GA lesion mask for previously unseen patients. By employing eye-specific latent vectors, we can generate accurate, individual predictions of lesion growth in shape. These predictions can be used to visualise how the disease is expected to evolve, potentially helping clinicians communicate future visual outcomes and treatment benefits to patients.

\section{Methodology}
Our framework uses an INR that models the disease trajectory as a continuous function $f_{\boldsymbol{\theta}}$ parameterised by $\boldsymbol{\theta}$. The INR consists of an auto-decoder, which is a multi-layer perceptron (MLP) with SIREN~\cite{sitzmannImplicitNeuralRepresentations2020} as the activation function. Following~\cite{danneckerPredictingLongitudinalBrain2026}, to capture individual changes in a setting with multiple images per patient, we assign one latent vector to each individual eye shared across all its images. In this way, we force the model to learn eye-specific features rather than visit-specific features. The model is conditioned on the latent vectors through modulation layers~\cite{mehtaModulatedPeriodicActivations2021}, where each vector gets linearly mapped to scale $\boldsymbol{\phi} \in \mathbb{R}^H$ and shift $\boldsymbol{\psi} \in \mathbb{R}^H$ parameters, where $H$ is the hidden size of the network. Similar to~\cite{danneckerCINeMAConditionalImplicit2026,danneckerPredictingLongitudinalBrain2026}, $\omega_0$ is applied only to the scale parameter to let the shift represent temporal dynamics as low-level signals. To give the model temporal information, we condition the INR on the number of weeks elapsed since the baseline visit and the patient's age at that visit. The two conditioning variables are concatenated to the latent vector before modulation. 
Following~\cite{vyasFitPixelsGet2026}, the INR is split into a reconstruction head $f_{\theta}^\text{faf}$ and a segmentation head $f_{\theta}^\text{seg}$, which are jointly modelled by the same MLP up to the penultimate layer $L-1$. Different from~\cite{vyasFitPixelsGet2026}, which feeds the segmentation head only from the penultimate layer, we concatenate the features of the last two SIREN layers and pass them through a linear layer to predict two-class probabilities (GA vs.\ background). As modulation layers, they are conditioned on the eye-specific latent vector and the time variables, as are all layers in our architecture.

\subsection{Training and Test-Time Adaptation}
\paragraph{\textbf{Training}.} The training pipeline is depicted in Fig.~\ref{fig:model_architecture}. First, a subset of coordinates $\mathbf{x}=(x,y)\in \mathbf{X}_i \subset \mathbb{R}^2$ is drawn from the domain $I_i : \mathbf X_i \to \mathbb R$ of a FAF image at time $t$ (measured in weeks from baseline) of eye $i$ and is given as input to the INR. Then, a latent vector $z_i$ sampled from $\mathcal{N}(0, 10^{-2})$ gets assigned to eye $i$. Following~\cite{danneckerCINeMAConditionalImplicit2026}, we make use of 3D spatial latent vectors $z_i\in\mathbb{R}^{C\times X_1\times X_2}$, where $C$ is the number of channels, and $X_1$ and $X_2$ the spatial dimensions. The value of the latent vector at $\mathbf{x}$ is obtained through bilinear interpolation, denoted as $\text{BiInterp}(z_i, \mathbf{x})$. The value of $t$ and the age of the patient at time $t$, $\text{age}_t$, are concatenated to the latent vector, resulting in $z_i(\mathbf{x}) = \left[\text{BiInterp}(z_i, \mathbf{x}) \oplus t \oplus \text{age}_t\right]$ which is then linearly mapped into scale and shift parameters in each modulation layer of the INR. The model parameters $\theta$ and the set of latent vectors $\{z_i\}_{i=1}^N$ are optimised to maximise the joint log posterior distribution over the $N$ eyes. This is achieved by minimising a combination of reconstruction loss $\mathcal{L}_{\text{MSE}}$, weighted by $\alpha$, and segmentation loss $\mathcal{L}_{\text{SEG}}$:
\begin{equation}
 \mathcal{L} = \alpha\mathcal{L}_{\text{MSE}}\!\left(
    f_{\theta}^{\text{faf}}\bigl(z_i(\mathbf{x})\bigr),\,
    I_i(\mathbf{x})
  \right)
  + \mathcal{L}_{\text{SEG}}\!\left(
    f_{\theta}^{\text{seg}}\bigl(z_i(\mathbf{x})\bigr),\,
    S_i(\mathbf{x})
\right)
\end{equation}
where $I_i(\mathbf{x})$ is the true pixel intensity value of the FAF image at $\mathbf{x}$, $S_i(\mathbf{x})$ is the true segmentation label at $\mathbf{x}$, and $f_{\theta}^{\text{faf}}$ and $f_{\theta}^{\text{seg}}$ are the corresponding predicted intensity value and segmentation label, respectively. $\mathcal{L}_{\text{SEG}}$ is the sum of Binary Cross Entropy loss and DICE loss.
\paragraph{\textbf{Test-Time Adaptation}.} During test-time adaptation, for a previously unseen eye $k$ a new latent vector $z_k$ is randomly sampled from $\mathcal{N}(0, 10^{-2})$ and assigned to it. While keeping $\theta$ frozen, the latent vector $z$ is optimised on the image modalities from $N-1$ of the $N$ available visits, using the same loss as for training. To predict the held-out images, the concatenation of optimised latent vector and conditions, $\Tilde{z}_k = \left[z_k \oplus t \oplus \text{age}_t\right]$, is then fed into the frozen INR, and a single forward pass is performed as
$
    f_{\theta}(\mathbf{x}|\Tilde{z}_k) = \left(f_{\theta}^{\text{faf}}(\mathbf{x}, \Tilde{z}_k(\mathbf{x})), f_{\theta}^{\text{seg}}(\mathbf{x}, \Tilde{z}_k(\mathbf{x}))\right)$.

\section{Experimental Setup}
\paragraph{\textbf{Dataset}.}
The method was evaluated on the OMEGA dataset \cite{ansariOpticalCoherenceTomography2023,ansariEvaluatingProgressionRetinal2025,valmaggiaOpticalCoherenceTomography}, an in-house longitudinal cohort of patients with GA secondary to AMD. The OMEGA dataset comprises scans of 37 eyes (19 right, 18 left) from 30 patients, monitored over up to 4 visits (approximately 12 weeks apart). The patients' ages range from 66 to 90 years, with an average of 78.4 years. At each visit, short-wavelength FAF images were performed using a \emph{Heidelberg Spectralis} device (Heidelberg Engineering, Heidelberg, Germany), generating $768\times 768$ grayscale images centred on the macula with a $30^\circ$ field of view and a scale of $\sim0.01$ mm/px. GA lesions were semiautomatically annotated in the FAF using the RegionFinder software (Heidelberg Engineering, Heidelberg, Germany) by three graders individually. For each visit, we then aggregated the three segmentation masks into a final mask using majority voting. The FAF images were pre-registered intra-patient by aligning each follow-up visit of an eye to its corresponding baseline with an adapted version of RIFT~\cite{liRIFTMultiModalImage2020}. The pixel intensities were normalised between [0,1], whereas spatial coordinates $\mathbf{x}$ and conditioning variables were normalised between [-1,1]. We cropped out the black frames introduced by image registration and resized our images to $512 \times 512$ pixels. We further resized to $256 \times 256$ pixels when comparing our method to all other DL models. The data were split into training (26 eyes), validation (5 eyes), and test (6 eyes) sets, patient-wise; that is, both eyes of a single patient were always included in the same split. In the validation set, to leave out one visit for evaluation, we included only patients with data from at least 2 visits.
Ethical approval was obtained from the Ethics Committee of Northwestern and Central Switzerland (BASEC ID: 2019-02003), and the study was performed according to the Declaration of Helsinki. Study enrolment took place at the Eye Clinic of the University Hospital Basel between March 2021 and July 2022, and written informed consent was obtained from all participants.
\paragraph{\textbf{Implementation Details}.}
The hyperparameters were optimised on the validation set. The best-performing model was then tested on the test set. The best-configured INR consisted of 8 modulation layers. The hidden layer size was set to 384, and the 3D latent vectors had shape $[256,32,32]$. Similar to~\cite{stolt-ansoNISFNeuralImplicit2023}, the value of $\alpha$ for the reconstruction loss was set to 10. During training, the learning rate was set to $10^{-4}$ for the INR and $5\cdot 10^{-3}$ for the latent vector. Following~\cite{sitzmannImplicitNeuralRepresentations2020}, $\omega_0$ was set to 30 for all layers. The network parameters and the latent vectors are jointly optimised using AdamW~\cite{loshchilovDecoupledWeightDecay2019b}, with a weight decay of $0.1$. The number of coordinates sampled at each iteration was set to 10'000. We trained and validated our model on a single NVIDIA RTX 2080 Ti 12GB for 50 epochs, taking on average $3.5$ hours and using $1.3$ GB of GPU memory. Test-time adaptation on a new eye took on average 1 minute with 25 epochs. 
\paragraph{\textbf{Comparison Methods}.}
For future FAF image prediction, we compare our method with ImageFlowNet~\cite{liuImageFlowNetForecastingMultiscale2025d}. ImageFlowNet operates on pairs of images from the same patient at two different time points $t_i$ and $t_k$, with $i<k$. By integrating a learned flow field, it predicts the FAF image at time $t_k$ using the image at time $t_i$ as input, along with the time shift $t_k - t_i$. To replicate this setup, which we refer to as \emph{Scenario~1}, we created all pairs of visits within each eye, optimised the latent vectors on the older visit, and tested them on the newer visit of the same pair. It also allows test-time optimisation by fine-tuning the flow on the full patient history to extrapolate the last image, which we refer to as \emph{Scenario~2} and represents the default test-time optimisation of our model. Similar to~\cite{liuImageFlowNetForecastingMultiscale2025d}, we include in the comparison a time-aware diffusion model (T-I2SBUNet) based on~\cite{liuI$^2$SBImagetoImageSchrodinger2023a}, and a time-conditional UNet (T-UNet) architecture derived from~\cite{hoDenoisingDiffusionProbabilistic2020}. We also include classical methods for comparison, such as linear interpolation~\cite{kayResultsLinearInterpolation1983}, cubic B-spline interpolation~\cite{houCubicSplinesImage1978}, and copy-forward, which simply copies the last available image as a prediction (no progression scenario).

\section{Results and Discussion} 
\begin{figure}[p]
\centering
\includegraphics[width=\textwidth]{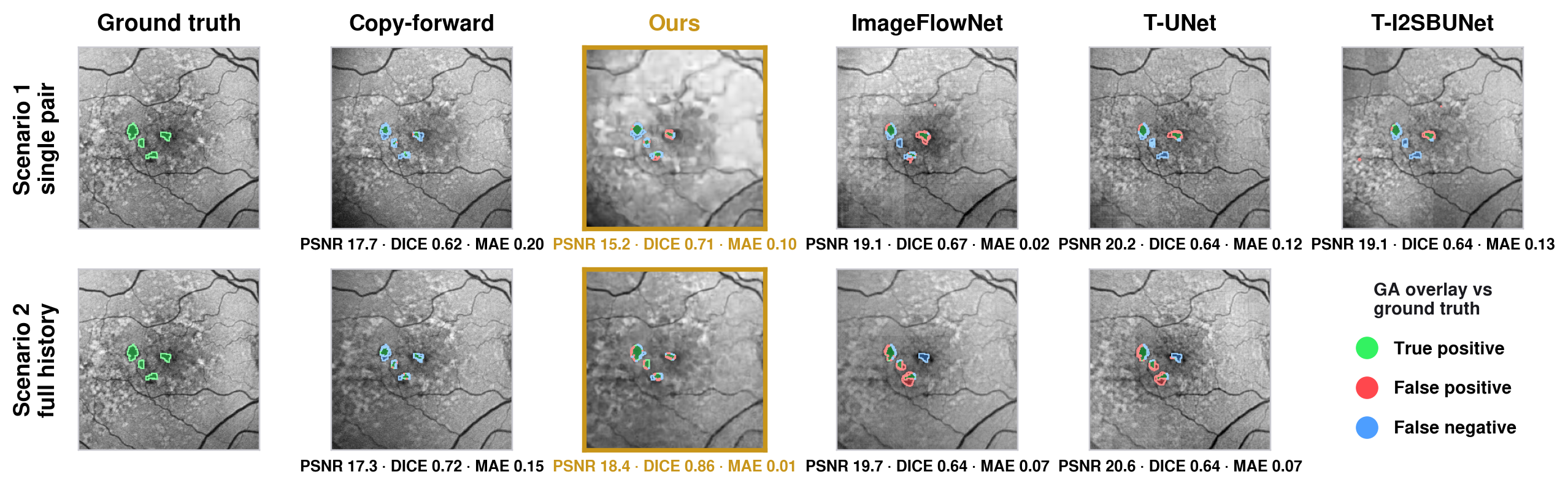}
\caption{Qualitative comparison between all different methods. In the first row, we show an example of \emph{Scenario 1}; in the second row, an example of \emph{Scenario 2} for the same test eye. PSNR, DICE score, and MAE are reported for each prediction. 
} 
\label{fig:scenario1_and_2}
\end{figure}

For a quantitative evaluation, we used Peak Signal-to-Noise Ratio (PSNR, in $\text{dB}$), Structural Similarity Index Measure (SSIM), and Learned Perceptual Image Patch Similarity (LPIPS) for assessing image reconstruction quality, and Hausdorff Distance (HD, in pixels), DICE score, and Mean Absolute Error (MAE, in $\text{mm}^2$) for segmentation quality. For a qualitative evaluation, we visually inspected the reconstructed FAF image quality and the segmented GA regions to assess their clinical accuracy. 
 \begin{table}[t]                                                                    
  \caption{Comparison against ImageFlowNet~\cite{liuImageFlowNetForecastingMultiscale2025d}, T-UNet~\cite{hoDenoisingDiffusionProbabilistic2020}, and T-I2SBUNet~\cite{liuI$^2$SBImagetoImageSchrodinger2023a} under the three prediction scenarios. PSNR/SSIM/LPIPS as FAF image reconstruction quality metrics; DICE/HD/MAE as GA segmentation quality metrics. Metrics are reported as $\text{mean}\pm\text{std}$. Best score per scenario highlighted in \textbf{bold}. Interp.: interpolation; Extrap.: extrapolation.}
  \label{tab:quantitativeresults}                                                                
  \centering                                                         
  \setlength{\tabcolsep}{2pt}                                                              
  \scalebox{0.8}{                                                                    
  \begin{tabular}{l|c|c|c|c|c|c}                                                                
      \toprule                                                                      
      Method                         
        & PSNR $\uparrow$ & SSIM $\uparrow$ & LPIPS $\downarrow$                                      
        & DICE $\uparrow$                      
        & HD $\downarrow$
        & MAE $\downarrow$ \\   
      \midrule                                                          
      \multicolumn{7}{l}{\textbf{\textsc{Extrapolation}} \normalfont(predict the last/future visit)}\\ 
      \midrule
      \multicolumn{7}{l}{\quad\textit{\textbf{Scenario 1}: single image pair (older $\rightarrow$ newer)}}\\    
      Copy-forward                                                                         
        & $\mathrm{16.49} \pm \mathrm{2.50}$ & $\mathrm{0.57} \pm \mathrm{0.08}$ & $\mathrm{\textbf{0.16}} \pm \mathrm{0.05}$ & $\mathrm{0.81} \pm \mathrm{0.15}$ & $\mathrm{\textbf{8.60}} \pm \mathrm{3.40}$ & $\mathrm{0.53} \pm \mathrm{0.47}$\\
      ImageFlowNet~\cite{liuImageFlowNetForecastingMultiscale2025d} 
        & $\mathrm{\textbf{17.99}} \pm \mathrm{2.12}$ & $\mathrm{0.58} \pm \mathrm{0.06}$ & $\mathrm{0.19} \pm \mathrm{0.03}$ & $\mathrm{0.65} \pm \mathrm{0.14}$ & $\mathrm{29.18} \pm \mathrm{16.71}$ & $\mathrm{1.29} \pm \mathrm{1.24}$ \\ 
      T-UNet~\cite{hoDenoisingDiffusionProbabilistic2020} 
        & $\mathrm{17.93} \pm \mathrm{1.99}$ & $\mathrm{0.59} \pm \mathrm{0.06}$ & $\mathrm{0.17} \pm \mathrm{0.03}$ & $\mathrm{0.71} \pm \mathrm{0.17}$ & $\mathrm{24.10} \pm \mathrm{15.92}$ & $\mathrm{0.82} \pm \mathrm{0.75}$ \\                                                                      
      T-I2SBUNet~\cite{liuI$^2$SBImagetoImageSchrodinger2023a}                                                            
        & $\mathrm{17.31} \pm \mathrm{2.67}$ & $\mathrm{0.60} \pm \mathrm{0.08}$ & $\mathrm{0.17} \pm 0.05$ & $\mathrm{0.77} \pm \mathrm{0.13}$ & $\mathrm{29.09} \pm \mathrm{21.03}$ & $\mathrm{0.56} \pm \mathrm{0.51}$ \\
      \textbf{Ours}                                                                 
        & $\mathrm{16.19} \pm \mathrm{0.51}$ & $\mathrm{\textbf{0.61}} \pm \mathrm{0.03}$ & $\mathrm{0.28} \pm \mathrm{0.03}$ & $\mathrm{\textbf{0.85}} \pm \mathrm{0.10}$ & $\mathrm{13.36} \pm \mathrm{8.57}$ & $\mathrm{\textbf{0.31}} \pm \mathrm{0.24}$ \\ 
      \addlinespace                                                             
      \multicolumn{7}{l}{\quad\textit{\textbf{Scenario 2}: test-time adaptation (full patient history $\rightarrow$ last)}}\\
      Copy-forward                                                                        
        & $\mathrm{\textbf{17.86}} \pm \mathrm{1.74}$ & $\mathrm{\textbf{0.61}} \pm \mathrm{0.09}$ & $\mathrm{\textbf{0.14}} \pm \mathrm{0.03}$ & $\mathrm{0.86} \pm \mathrm{0.09}$ & $\mathrm{6.98} \pm \mathrm{2.34}$ & $\mathrm{0.38} \pm \mathrm{0.24}$\\
      Linear extrap.~\cite{kayResultsLinearInterpolation1983}                                                
        & $\mathrm{12.62} \pm \mathrm{1.38}$ & $\mathrm{0.38} \pm \mathrm{0.12}$ & $\mathrm{0.32} \pm \mathrm{0.07}$ & $\mathrm{0.86} \pm \mathrm{0.09}$ & $\mathrm{6.98} \pm \mathrm{2.34}$ & $\mathrm{0.38} \pm \mathrm{0.24}$ \\   
      Cubic B-spline extrap.~\cite{houCubicSplinesImage1978} 
        & $\hphantom{0}\mathrm{7.31} \pm \mathrm{0.91}$ & $\mathrm{0.10} \pm \mathrm{0.03}$ & $\mathrm{0.74} \pm \mathrm{0.05}$ & $\mathrm{0.81} \pm \mathrm{0.11}$ & $\mathrm{7.19} \pm \mathrm{2.18}$ & $\mathrm{0.60} \pm \mathrm{0.40}$ \\
      ImageFlowNet~\cite{liuImageFlowNetForecastingMultiscale2025d}                
        & $\mathrm{14.18} \pm \mathrm{2.56}$ & $\mathrm{0.48} \pm \mathrm{0.07}$ & $\mathrm{0.29} \pm \mathrm{0.04}$ & $\mathrm{0.59} \pm \mathrm{0.24}$ & $\mathrm{69.92} \pm \mathrm{33.16}$ & $\mathrm{1.33} \pm \mathrm{1.65}$ \\    
      T-UNet~\cite{hoDenoisingDiffusionProbabilistic2020}
        & $\mathrm{17.81} \pm \mathrm{1.93}$ & $\mathrm{0.56} \pm \mathrm{0.10}$ & $\mathrm{0.27} \pm \mathrm{0.09}$ & $\mathrm{0.63} \pm \mathrm{0.19}$ & $\mathrm{33.30} \pm \mathrm{20.28}$ & $\mathrm{1.37} \pm \mathrm{1.16}$ \\ 
      \textbf{Ours}                                                                                  
        & $\mathrm{16.31} \pm \mathrm{3.01}$ & $\mathrm{\textbf{0.61}} \pm \mathrm{\textbf{0.04}}$ & $\mathrm{0.26} \pm \mathrm{0.04}$ & $\mathrm{\textbf{0.91}} \pm \mathrm{0.05}$ & $\mathrm{\textbf{6.78}} \pm \mathrm{2.58}$ & $\mathrm{\textbf{0.20}} \pm \mathrm{0.29}$ \\                                                
      \midrule                                                                                        
      \multicolumn{7}{l}{\textbf{\textsc{Missing visits}} \normalfont(reconstruct a held-out interior visit)}\\
      \midrule
      Linear interp.~\cite{kayResultsLinearInterpolation1983}                                                
        & $\mathrm{16.24} \pm \mathrm{2.46}$ & $\mathrm{0.45} \pm \mathrm{0.06}$ & $\mathrm{\textbf{0.23}} \pm \mathrm{0.10}$ & $\mathrm{0.85} \pm \mathrm{0.11}$ & $\mathrm{\textbf{14.46}} \pm \mathrm{5.06}$ & $\mathrm{0.29} \pm \mathrm{0.25}$ \\   
      Cubic B-spline interp.~\cite{houCubicSplinesImage1978} 
        & $\mathrm{14.44} \pm \mathrm{2.75}$ & $\mathrm{0.35} \pm \mathrm{0.06}$ & $\mathrm{0.34} \pm \mathrm{0.26}$ & $\mathrm{0.84} \pm \mathrm{0.12}$ & $\mathrm{14.64} \pm \mathrm{6.32}$ & $\mathrm{0.30} \pm \mathrm{0.25}$ \\ 
      \textbf{Ours}                               
        & $\mathrm{\textbf{17.88}} \pm \mathrm{1.30}$ & $\mathrm{\textbf{0.61}} \pm \mathrm{0.04}$ & $\mathrm{0.43} \pm \mathrm{0.08}$ & $\mathrm{\textbf{0.87}} \pm \mathrm{0.09}$ & $\mathrm{14.52} \pm \mathrm{5.27}$ & $\mathrm{\textbf{0.15}} \pm \mathrm{0.13}$ \\                    
      \bottomrule
  \end{tabular}}                                                                            
  \end{table}
\paragraph{\textbf{Quantitative and Qualitative Results.}}
 The quantitative results are presented in Table~\ref{tab:quantitativeresults}. In \emph{Scenario~1}, our model achieved performance comparable to both classical and DL approaches. Specifically, our method achieved among the best results in both segmentation quality (DICE: $0.85$; lesion-area MAE: $0.31$) and structural similarity (SSIM: $0.61$). In \emph{Scenario~2}, our method again achieved the best results across all segmentation metrics (DICE: $0.91$, HD: $6.78$; lesion-area MAE: $0.20\,\text{mm}^2$), demonstrating strong performance in learning disease progression in terms of lesion shape, localisation, and estimated area. We also achieved comparable performance on the FAF reconstruction task, demonstrating the model's ability to capture meaningful anatomical changes across both modalities. However, simply copying the last available ground truth yielded the best FAF reconstruction quality, which might be due to the short time shift between visits in our dataset.
 In Fig.~\ref{fig:scenario1_and_2}, an example of~\emph{Scenario~1} and \emph{Scenario~2} outputs on the same subject from all models is provided. In~\emph{Scenario~1}, our model achieved performance competitive with methods that were specifically trained on this setting. Our framework achieved the best results for lesion area prediction and GA segmentation, demonstrating the strength of having a single latent vector per eye, but at the cost of FAF reconstruction quality. In \emph{Scenario~2}, performance on both tasks improved significantly, yielding a much better FAF image and very accurate segmentation estimates. Notably, some methods (e.g.\ T-UNet, ImageFlowNet) achieved low lesion-area MAE yet predicted atrophic regions that visually mismatch the ground truth.
\begin{figure}[p]
\centering
\includegraphics[width=\textwidth]{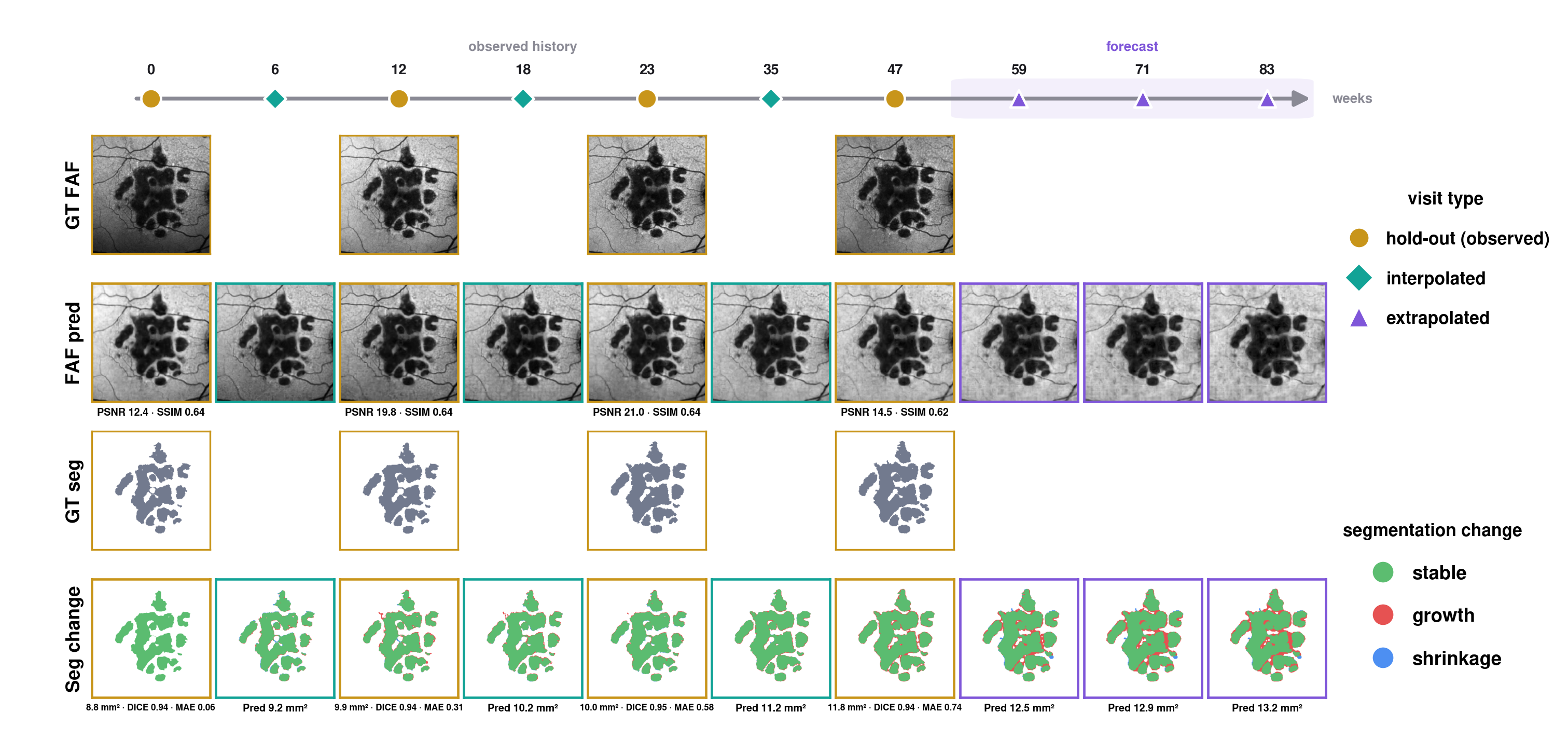}
\caption{Individual longitudinal prediction of a subject of the test set. First and third row: the four ground truth (GT) FAF and GA segmentation masks are shown. Second row: reconstructed existing visits (yellow), predicted interpolated (light blue) and extrapolated (purple) FAF images are provided. The last row displays the change maps for each predicted segmentation at time $t$ with respect to the last available GT.
}
\label{fig:longitudinal_predictions_mymodel}
\end{figure}
\begin{figure}
\centering
\includegraphics[width=\textwidth]{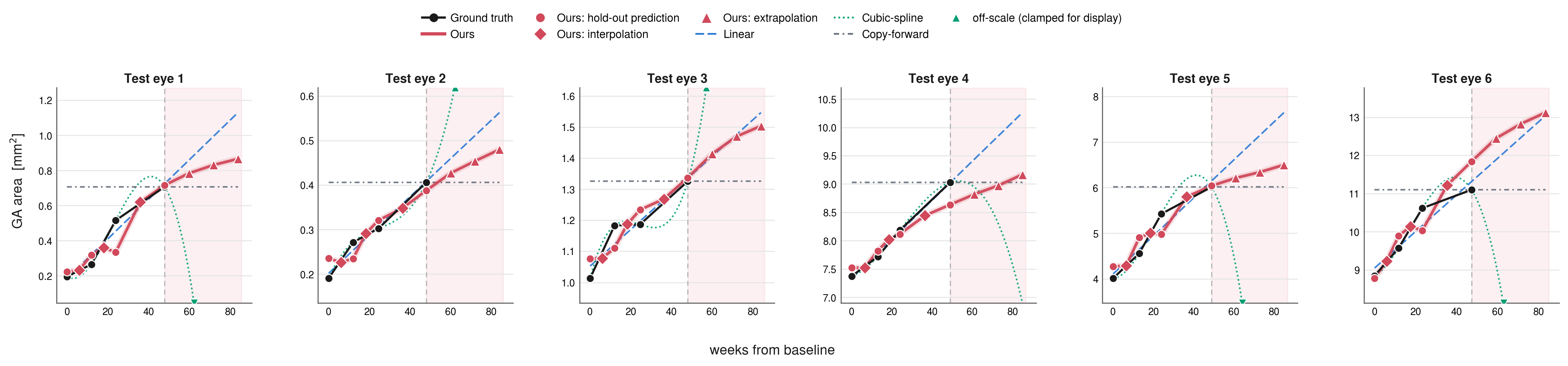}
\caption{Longitudinal analysis of the individual lesion area trajectories. The six subject-specific predicted trajectories are compared with linear extrapolation, cubic B-spline extrapolation, and copy-forward. Linear and cubic B-spline extrapolation curves are fitted using all ground truth visits.} 
\label{fig:lesion_size_trajectories}
\end{figure}
We also tested our model's ability to predict missing data by excluding data from one intermediate visit during test-time optimisation, and using the optimised latent vector to predict it. Here, our method achieved the best PSNR ($17.88$), SSIM ($0.61$), DICE ($0.87$), and MAE for the diseased area ($0.15$) compared to classical approaches such as linear and cubic B-spline interpolation. 
\paragraph{\textbf{Individual Disease Trajectories.}}
In Fig.~\ref{fig:longitudinal_predictions_mymodel}, the longitudinal predictions for a test subject are shown. Using the whole history of the subject, the model generated faithful predictions at new intermediate and future time points. The GA lesion areas extracted from the predicted interpolated and extrapolated segmentations are realistic relative to the ground truth values of nearby measurements and appear to follow the underlying disease progression curve. As shown in Fig.~\ref{fig:lesion_size_trajectories}, this is confirmed for each of the six eyes included in the test set, showing the subject-specific prediction capabilities of the method. Classical methods like linear extrapolation are still quite good at approximating the trajectory over time, mainly due to the limited sample size of the test set and the quite short observational period of the study, but looking at both interpolated and extrapolated values, on average, our method is able to yield the lowest lesion-area MAE even in such a scenario of limited data availability.
\begin{figure}[t]
\centering
\includegraphics[width=\textwidth]{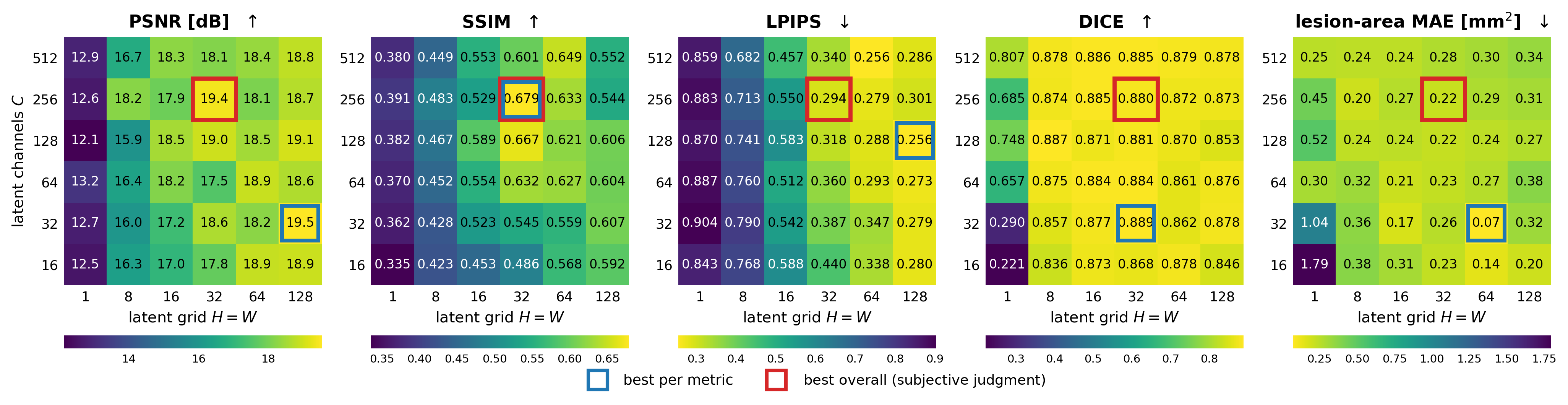}
\caption{Latent-grid ablation over number of channels $C$ (rows) and spatial resolution $H\!=\!W$ (columns). Each cell reports a metric at the checkpoint with the best DICE\,+\,lesion-area-MAE trade-off (\emph{lighter is better}). The red box is the best configuration ($C{=}256$, $H=W=32$) by subjective judgment over all five metrics.} 
\label{fig:latent_grid}
\end{figure}
\paragraph{\textbf{Hyperparameter Selection.}} We performed different stages of experiments. First, we varied how temporal and clinical information are combined in the decoder. We noticed that encoding time as an input coordinate, similar to~\cite{biederModelingNeonatalBrain2025}, consistently underperformed FiLM modulation. We therefore adopted raw-scalar modulation of (age, weeks) for all subsequent ablations. We then performed a grid search over spatial dimensions and the number of latent vector channels $C$. As shown in Fig.~\ref{fig:latent_grid}, DICE was largely insensitive to these choices. For both FAF reconstruction and segmentation, a $1 \times 1$ latent vector consistently underperformed larger latent grids, regardless of $C$. Larger grids led to better image reconstruction quality but at the cost of worse segmentation performance. In terms of lesion-area MAE, most configurations with spatial 2D latent vectors performed well, with intermediate latent grids yielding the best scores. Since no combination outperformed the others, we opted for a $32\times32$ grid and $C=256$ as a good trade-off between image reconstruction and segmentation quality.

\section{Conclusion}
We have introduced the first application of INRs for modelling the progression of GA secondary to AMD. We achieve competitive results in segmentation quality and in predicted changes in lesion shape and area. We demonstrate the usefulness of spatial latent vectors for learning the anatomical structure of each individual eye and for injecting time information via modulation layers. Our model also performs well at FAF prediction, both for past and future time points, and at predicting missing visits. The image quality of the predictions, however, especially during extrapolation, remains a challenge, and reconstructing the high-frequency details of FAF images will require further experiments and ablation studies. In some scenarios, simply copying the input image as the prediction yields the best image reconstruction quality across all comparison methods, primarily due to the study's short follow-up period. To this end, an evaluation criterion that incorporates additional anatomical information, such as GA lesion segmentation masks, as well as extracted biomarkers, is essential for assessing the clinical usefulness of the predictions and the individual disease evolution.
In future work, we would like to apply the method to a larger cohort of patients, including subjects at different stages of AMD, to generate risk prediction curves at the individual level and to analyse the latent vector space for more interpretable outcomes. We would also like to extend the method to multiple modalities, such as scanner laser ophthalmoscopy, and to apply it to 3D OCT volumes. 

    

\begin{credits}
\subsubsection{\ackname} The authors wish to thank all study participants. The data acquisition was funded by Boehringer Ingelheim. The work for this study was further supported by a grant of the Swiss Academy of Medical Sciences (YTCR 43/23).

\subsubsection{\discintname}
The authors have no competing interests to declare that are relevant to the content of this article.
\end{credits}

%
%
%
\bibliographystyle{splncs04}
\bibliography{mybibliography}
\end{document}